%% file: main.tex
\pdfoutput=1
\documentclass{article}
\usepackage{spconf,amsmath,graphicx}

\usepackage{mymacros}

\title{Grounded Language Understanding for Manipulation Instructions\\
Using GAN-Based Classification}
\name{Komei Sugiura and Hisashi Kawai\thanks{This work was partially supported by JST CREST and JSPS KAKENHI Grant Number 15K16074. The authors thank Dr. Peng Shen for his suggestions.
}}
\address{National Institute of Information and Communications Technology, Japan}

\begin{document}
\maketitle
\begin{abstract}
The target task of this study is grounded language understanding for domestic service robots (DSRs). In particular, we focus on instruction understanding for short sentences where verbs are missing. This task is of critical importance to build communicative DSRs because manipulation is essential for DSRs. Existing instruction understanding methods usually estimate missing information only from non-grounded knowledge; therefore, whether the predicted action is physically executable or not was unclear.

In this paper, we present a grounded instruction understanding method to estimate appropriate objects given an instruction and situation. We extend the Generative Adversarial Nets (GAN) and build a GAN-based classifier using latent representations. To quantitatively evaluate the proposed method, we have developed a data set based on the standard data set used for Visual QA. Experimental results have shown that the proposed method gives the better result than baseline methods.
\end{abstract}

\input{section1}
\input{section2}

\input{section3}
\input{section4}

\input{section6}

\input{section7}

\input{section8}


{\small
\bibliographystyle{IEEEbib}
\bibliography{reference}
}

\end{document}

%% file: section1.tex
\section{Introduction
\label{intro}
}

Based on increasing demands to improve the quality of life of those who need support, many DSRs are being developed\cite{I15}. The target task of this study is grounded language understanding for DSRs. In particular, we focus on instruction understanding for short sentences where verbs are missing. This task is of critical importance to build communicative DSRs because manipulation is essential for DSRs.

\begin{figure}[h]
 \centering
 \includegraphics[bb=0 0 439 323, height=42mm, width=58mm]{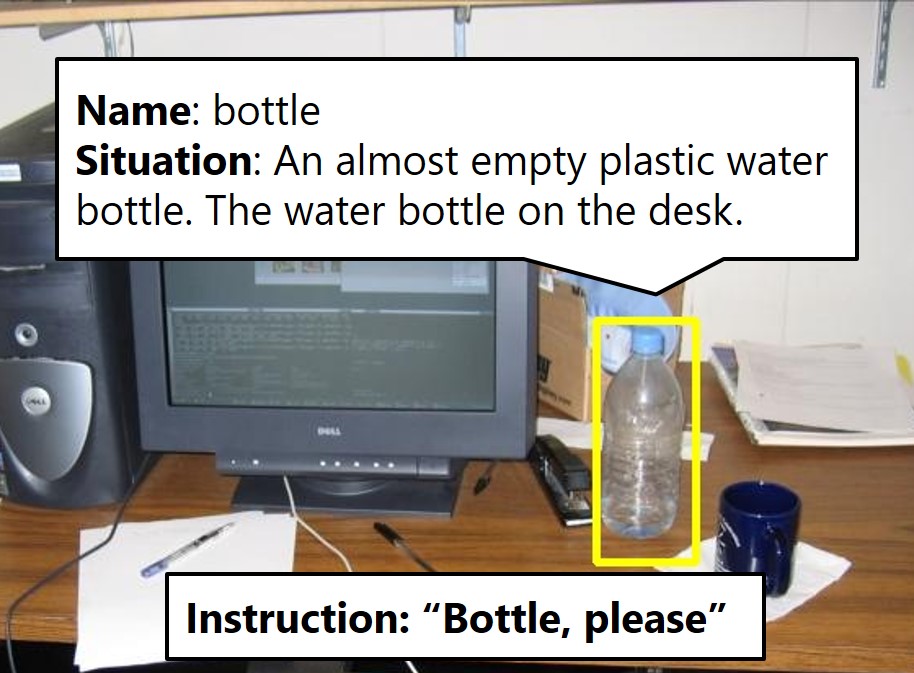}
 \hspace{1mm}
 \includegraphics[bb=0 0 162 329, height=42mm, width=22mm]{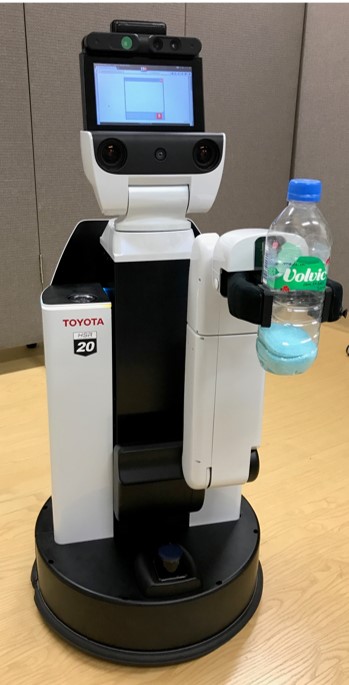}
 \caption{Left: Sample input used in the experiment. Verb is missing in the instruction. Right: Standard domestic service robot platform\cite{H14}.}
 \label{eyecatch}
\end{figure}

An example situation where a user asks a DSR to fetch a bottle is shown in the left-hand figure of \figref{eyecatch}. The right-hand figure of \figref{eyecatch} shows a standard DSR platform.

Existing instruction understanding methods usually estimate missing information only from non-grounded knowledge; therefore, whether the predicted action is physically executable or not is unclear. Moreover, the interaction sometimes takes more than one minute until the robot starts to execute the task in a typical setting. Such interactions are inconvenient for the user.

In this paper, we present a grounded instruction understanding method to estimate appropriate objects given an instruction and situation. We extend the GAN\cite{G7} and build a GAN-based classifier using latent representations. 
Unlike other methods, the user does not need to directly specify which action to take in the utterance, because it is estimated. 

There have been many studies on the variations of GANs (e.g., \cite{mirza2014conditional,chen2016infogan}).
Recently, some studies applied GANs for classification tasks\cite{springenberg2015unsupervised, S21}. Our study is inspired by these method; however, the difference is that our method has Extractor. Extractor's task is to convert raw input to latent representations that are more informative for classification. Generator's task is data augmentation to improve generalization ability.

The following are our key contributions: 
\begin{itemize}
 \item We propose a novel GAN-based classifier called LAtent Classifier Generative Adversarial Nets (LAC-GAN). LAC-GAN is composed of three main components: Extractor $E$, Generator $G$, and Discriminator $D$. The method is explained in \secref{lacgan}.
 \item LAC-GAN is applied to manipulation instruction understanding. 
       To quantitatively evaluate the proposed method, we have developed a data set based on the standard data set used for Visual QA\cite{vinyals2015show}. The results are shown in Section 7.
\end{itemize}

%% file: section2.tex
\section{Related Work
\label{related}
}

The classic dialogue management mechanisms adopted for DSRs process linguistic and non-linguistic information separately. More recently, the robotics community has started to pay more attention to the mapping between language and real-world information, mainly focusing on motion\cite{K5,S1,I3}. Kollar {\it et al.} proposed a path planning method from natural language commands\cite{K14}. However, most of the SLU methods used for DSRs are still rule-based\cite{vanzo2016robust}. In the dialogue community, Lison {\it et al.} presented a model for priming speech recognition using visual and contextual information\cite{L13}.

We developed LCore\cite{sugiura09interspeech}, which is a multimodal SLU method. In LCore, the SLU was integrated with image, motion prediction, and object relationships. However, the grammar and vocabulary was limited and the situation was artificial. We also developed Rospeex\footnote{http://rospeex.org}, which is a multilingual spoken dialogue toolkit for robots\cite{sugiura15iros}. Rospeex has been used by 45,000 unique users.
In recent years, there have been many studies on image-based caption generation and visual QA\cite{vinyals2015show,K25}. Our study has a deep relationship with these studies, however the difference is that our focus is not on caption generation.

There have been many studies on the variations of GANs (e.g., \cite{G7,mirza2014conditional,chen2016infogan}).
A GAN is usually composed of two main components: Generator $G$ and Discriminator $D$.
AC-GAN\cite{odena2016conditional} uses category labels as well as the estimated source (real or fake) as Discriminator's output. Most of GAN-related studies focus on generating pseudo samples, e.g., image and text. 
Recently, some studies applied GANs for classification tasks\cite{springenberg2015unsupervised, S21}. Our study is inspired by these method; however, the difference is that LAC-GAN has Extractor. 



%% file: section3.tex
\section{Task Definition
\label{task}
}


In this paper, we focus on grounded language understanding for manipulation instructions. This problem is of critical importance to build communicative DSRs because manipulation is essential for DSRs. In particular, we focus on instruction understanding for short sentences missing verbs.

Specifically, the following is a typical use case considered in this study:
\begin{quote}
 U: ``Robot, bottle please.''\\
 R: ``Please select from the list (GUI shows a list of manipulable bottles).''
\end{quote}
The task here is to estimate whether candidate objects are likely to be manipulable.
This is challenging because the physical situation should be modeled to understand the instruction.

On the other hand, most DSRs use non-grounded language understanding to solve such a task. Missing information is estimated only from linguistic knowledge; therefore, whether the predicted action is physically executable or not is unclear. Moreover, the interaction sometimes takes more than one minute until the robot starts to execute the task in a typical setting\cite{I15}. Such interactions are inconvenient for the user.


Here, we define terminology used in this study as follows:
\begin{itemize}
 \item An {\it manipulation instruction understanding} is defined to be classifying the target object (trajector) as manipulable or not given the situation.
 \item The {\it situation} is defined as a set of sentences explaining a (camera) image.
 \item A {\it trajector} is the target object which is focused in the scene\cite{L8}.
 \item The trajector is {\it manipulable} if a typical DSR is technically capable of manipulating it given the situation.
\end{itemize}
The key evaluation metric in this study is the classification accuracy.


%% file: section4.tex
\section{Generative Adversarial Nets
\label{gan}
}

GAN\cite{G7} is composed of two main components: Generator $G$ and Discriminator $D$.
Input to $G$ is a $d_z$-dimensional random variable, $\vec{x}$. Output from $G$ is $ \vec{x}_{fake}$ defined as follows:
\begin{align}
 \vec{x}_{fake} = G(\vec{z}).
\end{align}

The input source of $D$ is denoted as $S$, which is selected from the set $\{real, fake\}$.
When $S$ is $real$, a real training sample denoted as $\vec{x}_{real}$ is input to $D$.
$D$'s task is to discriminate the source, $S \in \{real, fake\}$, given $\vec{x} \in \{\vec{x}_{real}, \vec{x}_{fake}\}$. On the other hand, $G$'s task to fake $D$.

$D$ outputs the likelihood of $S$ being $real$ given $\vec{x}$ as follows:
\begin{align}
 D(\vec{x}) &= p(S = real | \vec{x}).
\end{align}

The following cost functions are used to optimize GAN's network parameters.
\begin{align*}
 J^{(D)} &= - \frac{1}{2} \mathbb{E}_{\vec{x}_{real}} \log D(\vec{x}_{real})
 - \frac{1}{2} \mathbb{E}_{\vec{z}} \log (1-D(G(\vec{z}))),\\
 J^{(G)} &= - J^{(D)},
\end{align*}
where $J^{(D)}$ and $J^{(G)}$ denote the cost functions of $D$ and $G$, respectively. In the training process, the training for $D$ and $G$ are conducted alternately. First, $D$'s parameters are trained, and then $G$'s parameters are trained. $D$'s parameters are fixed while $G$'s parameters are trained.


\section{Latent Classifier GAN
\label{lac-gan}
}

\begin{figure*}[t]
 \centering
 \includegraphics[bb=0 0 685 263, height=50mm]{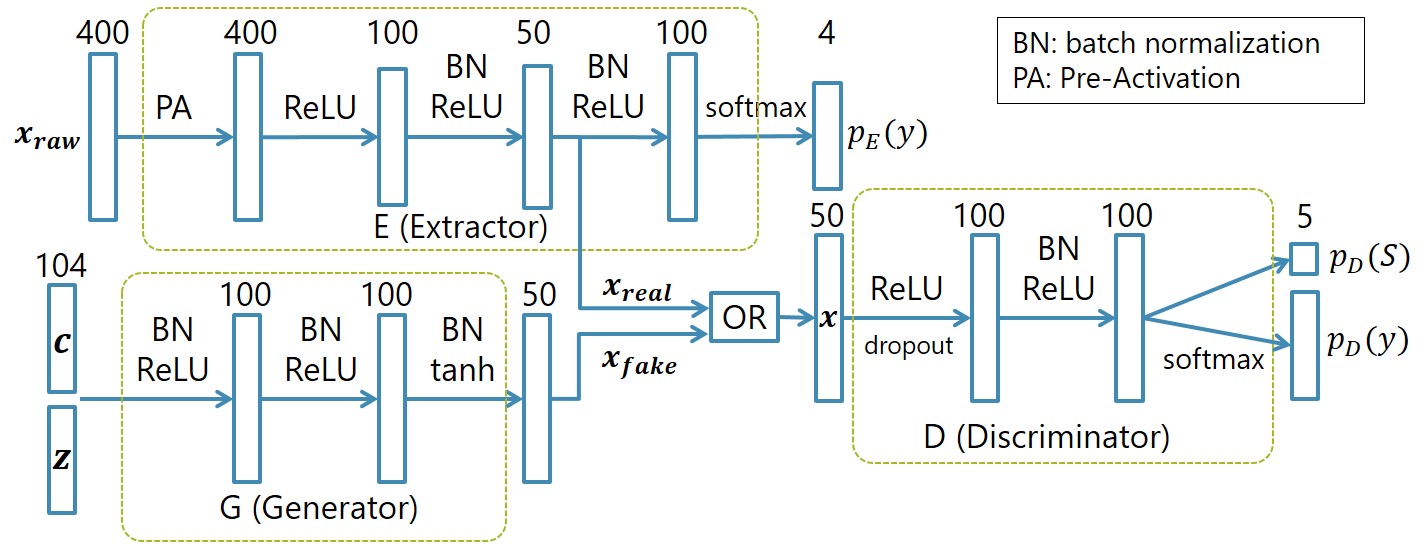}
 \caption{
 Model structure of LAC-GAN. The numbers on the layers represent the node numbers.
}
\label{lacgan}
\end{figure*}

\subsection{Model Structure}

We extend GAN for classification, and propose a novel method called LAtent Classifier Generative Adversarial Networks (LAC-GAN). Our approach is inspired by the fact that $G$ does not have to generate raw representations, e.g., image or text, in classification tasks. Instead in our approach, $G$ is used for data augmentation, and asked to generate latent representations of the data. LAC-GAN's model structure is shown in \figref{lacgan}. LAC-GAN is composed of three main components: Extractor $E$, Generator $G$, and Discriminator $D$.

Suppose we obtain a training sample $(\vec{x}_{raw}, y)$, where $\vec{x}_{raw}\in\mathbb{R}^{d_{raw}}$ and $y$ denote raw features and the label, respectively.
We assume that $y$ is a categorical variable, which is a $d_y$-dimensional binary vector.

Unlike other studies where GANs are used for sample generation, our focus in on GAN-based classification. From this background, it is reasonable to convert $D$'s input to more informative features in terms of classification. Such features are extracted by $E$ from $\vec{x}_{raw}$.

Input to $E$ is $\vec{x}_{raw}$, and output from $E$ is $p_E(y)$, which is the likelihood of $y$ given $\vec{x}_{raw}$. In the optimization process of $E$, we simply minimize the following cross-entropy-based cost function:
\begin{align}
 J_C = - \sum_j y_j \log p_E(y_j), \label{eq41}
\end{align}
where $y_j$ denotes the label for $j$-th category. We designed $E$'s structure as a bottleneck network, in which the output of the bottleneck layer is extracted as a $d_{real}$-dimensional vector, $\vec{x}_{real}$.

Input to $G$ is category denoted as $c$ and a $d_z$-dimensional random vector denoted as $\vec{z}$. 
In each mini-batch, new random samples are drawn as $c$ and $\vec{z}$ from a categorical distribution and a continuous distribution, respectively. The standard normal distribution was used as the continuous distribution: $\vec{z} \sim \mathcal N(\vec{0}, I)$. Output from $G$ is denoted as $\vec{x}_{fake}$, which is a $d_{fake}$-dimensional continuous vector. In LAC-GAN, $G$'s role is data augmentation; therefore, $G$ is expected to generate $\vec{x}_{real}$-like samples to improve generalization ability.

OR gate in \figref{lacgan} shows that either $\vec{x}_{real}$ or $\vec{x}_{fake}$ is input to $D$. 
$D$'s task is two-fold. The first one is to discriminate the source, $S \in \{real, fake\}$, given $\vec{x} \in \{\vec{x}_{real}, \vec{x}_{fake}\}$. The other is to categorize the input.
Therefore, $D$ has two types of output: the likelihood of $S$ given $\vec{x}_{real}$, $p_D(S)$, and the likelihood of $y$ given $\vec{x}_{real}$, $p_D(y)$.

Similar to other GAN-based classification models\cite{odena2016conditional, S21}, we define separate cost functions for $p_D(S)$ and $p_D(y)$. The former is defined as follows:
\begin{align}
 J_S &= - \frac{1}{2} \mathbb{E}_{\vec{x}_{real}} \log D(\vec{x}_{real}) \nonumber \\ 
 &\hspace{20mm}- \frac{1}{2} \mathbb{E}_{\vec{z}, c} \log (1-D(G(\vec{z}, c))),
\end{align}
where $G(\vec{z}, c)$ is the output of $G$ given $\vec{z}$ and $c$\cite{mirza2014conditional}.
The same cost function as \eqref{eq41} is used for the latter, where $p_E$ is rewritten as $p_D$.
The total cost function for $D$ is defined as the weighted sum of the two.
The weight parameter is denoted as $\lambda$.

Thus, the cost functions for LAC-GAN are defined as follows:
\begin{align}
 J^{(E)}_{lacgan} &= J_C,\\
 J^{(D)}_{lacgan} &= J_S + \lambda J_C,\\
 J^{(G)}_{lacgan} &= - J_S,
\end{align}
where $J^{(E)}_{lacgan}, J^{(D)}_{lacgan}$, and $J^{(E)}_{lacgan}$ denote the cost functions of $E, D$, and $G$.

\subsection{Activation Functions and Regularization}

We use batch normalization (BN\cite{I16}) to regularize the parameters of a layer. BN reduces internal covariate shift to stabilize the training by converting input mean 0 and variance 1 within each mini-batch.
Since BN act as a standardization method, dropout does not have to be used where BN is applied.
We do not apply BN in the first layer of $D$, which is standard in GAN-based approaches.
We use dropout instead for the layer.

BN is usually applied after addition, which is called post-activation. In pre-activation (PA), BN is applied before addition. PA outperformed post-activation in the CIFAR-10 task when the network is very deep\cite{H15}.
As explained in Section 5, the $\vec{x}_{raw}$ is represented as a paragraph vector\cite{L21}, which is not standardized. We apply PA to standardize the data within each mini-batch.

We use ReLU, softmax, $\tanh$ as activation functions. Since the output of the $E$ and $D$ is a categorical variable, we use softmax in their final layers. The output of $G$ is a positive/negative continuous values; therefore, we use $\tanh$ in $G$'s final layer. We use ReLU for the other layers. Leaky ReLU was reported to show better performance than ReLU; however, we did not obtain statistically significant results in pilot experiments for our task.


%% file: section6.tex
\section{Developing Object Manipulation Multimodal Data Set
\label{dataset}
}

As far as we know, no standard data set exists for object-manipulation instruction understanding. Therefore, we first explain the data set developed for this study. To avoid creating a data set that is too artificial, we extracted a subset from the standard Visual Genome dataset\cite{K25}.

The Visual Genome dataset contains over 100k images, where each image has as average of 21 objects. The bounding boxes of the objects are given by human annotators, and they are canonicalized to the WordNet synsets\cite{M17}. Each image also contains regions with descriptions given by human annotators. Unlike other datasets such as MS-COCO\cite{lin2014microsoft}, the Visual Genome dataset contains more bounding boxes and their descriptions per image. This is suitable for our problem setting because rich representation is available for a situation. Another advantage is that the data set contains a wide variety of images; therefore, we can empirically validate classification methods in various situations.

Next, we selected target synsets that were likely to be used in DSR use cases. In this paper, we extracted a bounding box as a sample if its label is either of the following synsets:
\begin{itemize}
 \item apple, ball, bottle, can, cellular telephone, cup, glass, paper, remote control, shoe, or teddy(bear),
\end{itemize}
where ``n.01'' is not written for readability. These synsets were randomly selected from the synsets that are often used in objection manipulation tasks by DSRs. The images were filtered by the above synsets, and they were also filtered by the minimum height and width of the objects. In this study, both are set to 50 pixels. The images were randomly extracted from the Visual Genome data set, and the number of images were balanced among the above synsets.

\begin{figure}[t]
 \centering
\begin{minipage}[c]{42mm}
 \includegraphics[bb=0 0 323 216,height=30mm]{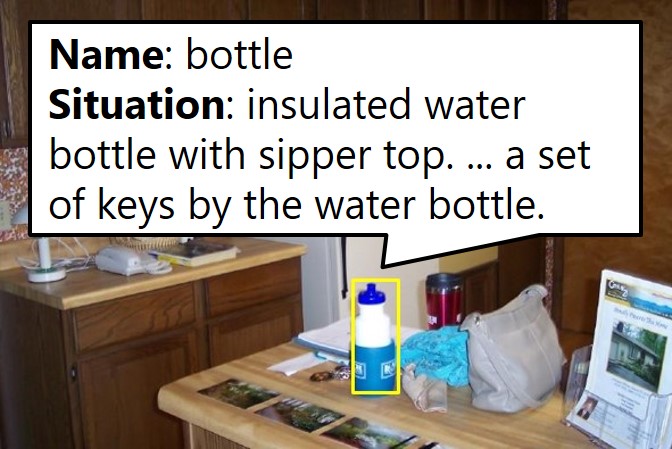}
\end{minipage}
\hspace{2mm}
\begin{minipage}[c]{38mm}
 \includegraphics[bb=0 0 288 216,height=30mm]{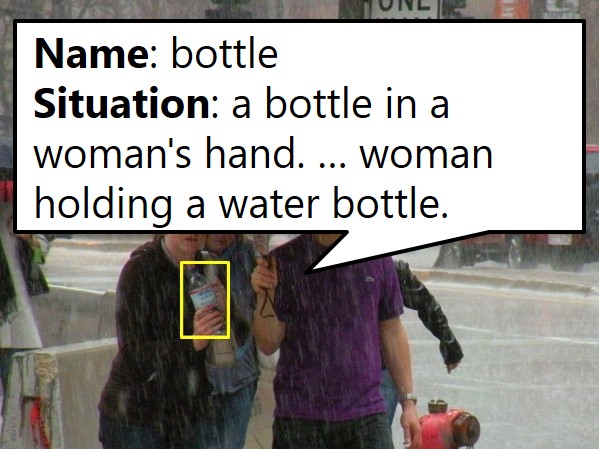}
\end{minipage}
 \caption{
 Two representative samples for the synset ``bottle.n.01''.
 Each yellow box shows the bounding box of the trajector.
 The left sample is labeled as ``positive'' because the trajector is manipulable under the situation.
 However, the right sample is labeled as ``negative'' because the trajector is already grasped and the robot cannot manipulate it.
}
\label{bottle}
\end{figure}

Next, the samples were labeled using the following criteria. 
\begin{itemize}
 \item[(E1)] The bounding box contains multiple objects of the same kind, e.g. several shoes in a basket.
 \item[(E2)] The bounding box does not contain necessary information about the object, e.g. the handle of a glass.
 \item[(N)] The bounding box sufficiently contains the trajector; however, the trajector is not suitable for grasping. For example, a meat ball is categorized as part of the synset ``ball.n.01'', however the robot should not grasp it.
 \item[(M0)] The object is not manipulable in that situation. In other words, the path planning for manipulation fails. This category includes cases where the trajector is surrounded by many obstacles, it is held by a human, or it is moving.
 \item[(M1)] The object is manipulable, however autonomous grasping could fail in the situation. If the robot is remotely controlled, the object can be safely grasped.
 \item[(M2)] The object is manipulable, and the robot can autonomously manipulate the object in that situation.
 \item[(O)] None of the above.
\end{itemize}
Examples of the images are shown in \figref{bottle}. The annotator was one of the authors and a robotics expert. To exclusively label a sample, the criteria were checked in the same order as the above list. For example, if the sample was labeled as (E1), it was never labeled as (M0).

In the experiments, the samples categorized as (N), (M0), (M1), and (M2) were used; therefore, the task is a 4-class classification problem. Categories (E1) and (E2) were not used because it is unlikely that sufficient situation information would be available. 

The data samples were shuffled and divided into the training (80\%), validation (10\%), and test (10\%) sets. The statistics of the original and labeled data set are shown in \tabref{dataset}. Hereafter, we call the labeled data set the ``Object Manipulation Multimodal Data Set''.


\begin{table}[b]
 \centering
 \caption{Statistics of the Object Manipulation Multimodal Data Set. The abbreviated categories are defined in Section 6.}
{\small
 \input{fig/dataset/dataset}
 \label{dataset}
}
\end{table}


%% file: fig/dataset/dataset.tex




\begin{tabular}{|l|c|}
\hline 
Data set size (all categories)& 896\tabularnewline
\hline 
Number of unique words describing situations & 7926\tabularnewline
\hline 
Average number of words describing situations & 305\tabularnewline
\hline 
Training-set size (N, M0, M1, M2) & 539 (80\%)\tabularnewline
\hline 
Validation-set size (N, M0, M1, M2) & 67 (10\%)\tabularnewline
\hline 
Test-set size (N, M0, M1, M2)& 67 (10\%)\tabularnewline
\hline 
\end{tabular}

%% file: section7.tex
\section{Experiments
\label{exps}
}

\subsection{Setup}

In the experiments, we assumed that the input was given as the linguistic expressions of an instruction and situation. Here, the instruction did not contain a verb, but contained the trajector's ID. The linguistic expressions were obtained from the Object Manipulation Multimodal Data Set. 


The input to LAC-GAN is given as follows:
\begin{align*}
 \vec{x}_{raw} = \{\vec{x}_{name}, \vec{x}_{situation}\},
\end{align*}
where $\vec{x}_{name}$ and $\vec{x}_{situation}$ denote the embedded representations of the trajector's name and the situation, respectively.

We used the distributed memory model of paragraph vector (PV-DM\cite{L21}) to obtain the embedded representations. First, the name expression was obtained based on the trajector's ID. Most of the name expressions contained only one candidate consisted of a noun; however, some expressions contained multiple candidates consisted of multiple words, e.g., ``cups in stack $\mid$ stacked cups.'' Regardless of the number of words, the name expressions were converted to a 200-dimensional paragraph vector, $\vec{x}_{name}$. This is done by averaging the paragraph vector of the candidates. The situation was composed of multiple descriptions of other objects in the scene. Those descriptions were converted to a 200-dimensional vector, $\vec{x}_{situation}$.

To train the PV-DM, we extracted descriptions from the Visual Genome data set, and built a corpus.
The corpus consisted of 4.72 million sentences, where the average length of a sentence was 5.18 words.

\tabref{settings} shows the experimental setup. The random variable $\vec{z}$ was sampled from $\mathcal{N}(\vec{0}, I)$, where $d_z$ was set to $d_z = 100$ ; however, preliminary experimental results showed that the effect of $d_z$ was not large among all hyper-parameters. The dimensions $d_{raw}, d_{real}, d_{fake}$ and $d_y$ were set to $d_{raw}=400$, $d_{real} = d_{fake} = 50$, and $d_y=4$, respectively.

\begin{table}[t]
 \centering
 \caption{Experimental setup. Extractor, Generator, and Discriminator are denoted as $E, G,$ and $D$, respectively.}
{\small
 \input{fig/settings/settings01en}
}
 \label{settings}
\end{table}

\subsection{Results}

We compared our method (LAC-GAN) with baseline methods including AC-GAN\cite{odena2016conditional} using the Object Manipulation Multimodal Data Set. In general in training deep networks, the accuracy does not monotonically increases as the increase in epochs. Due to the cost of cross-validation in deep networks, the best model is usually considered to be the model which gives the highest validation-set accuracy in the standard experimental protocol. According to this protocol, the test-set accuracy obtained by each best model was compared. The result is shown in \tabref{accuracy}.

To make the comparison fair, the proposed and baseline methods were made to have the same structure and the same number of nodes except the input layer. ``With/without PA'' represents whether the pre-activation was used or not. ``Extractor only'' represents the test-set accuracy based on the output of Extractor, $p_E(y)$; therefore, this means the test-set accuracy obtained by a simple six-layered feed-forward network.

\tabref{accuracy} shows that LAC-GAN outperformed the baseline methods including AC-GAN and ``Extractor only''. From the comparison between LAC-GAN and AC-GAN, it is indicated that we can obtain better performance by extracting informative features. From the comparison between LAC-GAN and ``Extractor only'', it is indicated that GAN-based data generation can improve the test-set accuracy. This also indicates that the generalization ability is enhanced by LAC-GAN.

\begin{table}[t]
 \centering
 \caption{Test-set accuracy obtained from the best models of each method. The best model is obtained as the model which gives the highest validation-set accuracy.}
{\small
 \input{fig/classification/lacgan01en}
}
 \label{accuracy}
\end{table}



%% file: fig/settings/settings01en.tex




\begin{tabular}{|c|c|}
\hline 
Optimization & Adam (Learning rate$=0.0005$,\tabularnewline
 method & $\beta_1=0.5$, $\beta_2=0.999$)\tabularnewline
\hline 
$d_{raw}$ & Name\,(200) + Situation\,(200)\tabularnewline
\hline 
Num. nodes ($E$) & 400\,(input), 400, 100, 50, 100, 4\,(output)\tabularnewline
\hline 
Num. nodes ($G$) & 104\,(input), 100, 100, 50\,(output)\tabularnewline
\hline 
Num. nodes ($D$) & 50\,(input), 100, 100, 5\,(output)\tabularnewline
\hline 
Batch size & 50\,($E$), 20\,($G$ and $D$)\tabularnewline
\hline 
Weight $\lambda$ & 0.2\tabularnewline
\hline 
\end{tabular}

%% file: fig/classification/lacgan01en.tex




\begin{tabular}{|c|c|}
\hline 
 Method & Test-set accuracy\tabularnewline
\hline 
\hline 
Baseline (AC-GAN\cite{L21}, without PA) & 50.7\%\tabularnewline
\hline 
Baseline (AC-GAN, with PA) & 58.2\%\tabularnewline
\hline 
Extractor only & 61.1\%\tabularnewline
\hline 
Ours (LAC-GAN) & 67.1\%\tabularnewline
\hline 
\end{tabular}

%% file: section8.tex
\section{Conclusion
\label{conclusion}
}

Based on increasing demands to improve the quality of life of those who need support, many DSRs are being developed. Although there are still many tasks that DSRs cannot do, they have advantages over human support staff and service dogs. A human carer cannot work without rest, and training a service dog requires nearly two years.

In this paper, we presented a grounded language understanding method to estimate manipulability from short instructions. We extended the GAN\cite{G7} to build LAC-GAN, which is a GAN-based classifier using latent representations. To quantitatively evaluate LAC-GAN, we have developed the Object Manipulation Multimodal Data Set. Linguistic expressions on the trajector and situation are extracted from the data set and converted into paragraph vectors by PV-DM\cite{L21}. The manipulability is predicted based on the paragraph vectors by LAC-GAN.
We experimentally validated LAC-GAN, and found it gives the better result than baseline methods including AC-GAN\cite{odena2016conditional}. Future directions include the integrating the proposed method with object detection and caption generation.



%% file: main.bbl
\begin{thebibliography}{10}

\bibitem{I15}
Luca Iocchi, Dirk Holz, Javier Ruiz-del Solar, Komei Sugiura, and Tijn van~der
  Zant,
\newblock ``{RoboCup@Home: Analysis and Results of Evolving Competitions for
  Domestic and Service Robots},''
\newblock {\em {Artificial Intelligence}}, vol. 229, pp. 258--281, 2015.

\bibitem{G7}
Ian Goodfellow, Jean Pouget-Abadie, Mehdi Mirza, Bing Xu, David Warde-Farley,
  Sherjil Ozair, Aaron Courville, and Yoshua Bengio,
\newblock ``{Generative Adversarial Nets},''
\newblock in {\em Advances in Neural Information Processing Systems}, 2014, pp.
  2672--2680.

\bibitem{mirza2014conditional}
Mehdi Mirza and Simon Osindero,
\newblock ``{Conditional Generative Adversarial Nets},'' arXiv preprint
  arXiv:1411.1784, 2014.

\bibitem{chen2016infogan}
Xi~Chen, Yan Duan, Rein Houthooft, John Schulman, Ilya Sutskever, and Pieter
  Abbeel,
\newblock ``{InfoGAN: Interpretable Representation Learning by Information
  Maximizing Generative Adversarial Nets},''
\newblock in {\em Advances in Neural Information Processing Systems}, 2016, pp.
  2172--2180.

\bibitem{springenberg2015unsupervised}
Jost~Tobias Springenberg,
\newblock ``{Unsupervised and Semi-Supervised Learning with Categorical
  Generative Adversarial Networks},'' arXiv preprint arXiv:1511.06390, 2015.

\bibitem{S21}
Peng Shen, Xugang Lu, Sheng Li, and Hisashi Kawai,
\newblock ``{Conditional Generative Adversarial Nets Classifier for Spoken
  Language Identification},'' Proc. of Interspeech, 2017.

\bibitem{vinyals2015show}
Oriol Vinyals, Alexander Toshev, Samy Bengio, and Dumitru Erhan,
\newblock ``Show and tell: A neural image caption generator,''
\newblock in {\em Proceedings of the IEEE Conference on Computer Vision and
  Pattern Recognition}, 2015, pp. 3156--3164.

\bibitem{H14}
Kunimatsu Hashimoto, Fuminori Saito, Takashi Yamamoto, and Koichi Ikeda,
\newblock ``A field study of the human support robot in the home environment,''
\newblock in {\em Advanced Robotics and its Social Impacts (ARSO), 2013 IEEE
  Workshop on}, 2013, pp. 143--150.

\bibitem{K5}
Volker Kr{\"u}ger, Danica Kragic, Ales Ude, and Christopher Geib,
\newblock ``{The Meaning of Action: A Review on Action Recognition and
  Mapping},''
\newblock {\em {Advanced Robotics}}, vol. 21, no. 13, pp. 1473--1501, 2007.

\bibitem{S1}
Yuuya Sugita and Jun Tani,
\newblock ``Learning semantic combinatoriality from the interaction between
  linguistic and behavioral processes,''
\newblock {\em Adaptive Behavior}, vol. 13, no. 1, pp. 33--52, 2005.

\bibitem{I3}
Tetsunari Inamura, Iwaki Toshima, Hiroaki Tanie, and Yoshihiko Nakamura,
\newblock ``{Embodied Symbol Emergence Based on Mimesis Theory},''
\newblock {\em {International Journal of Robotics Research}}, vol. 23, no. 4,
  pp. 363--377, 2004.

\bibitem{K14}
T.~Kollar, S.~Tellex, D.~Roy, and N.~Roy,
\newblock ``{Toward Understanding Natural Language Directions},''
\newblock in {\em Proceeding of the 5th ACM/IEEE international conference on
  Human-robot interaction}, 2010, pp. 259--266.

\bibitem{vanzo2016robust}
Andrea Vanzo, Danilo Croce, Emanuele Bastianelli, Roberto Basili, and Daniele
  Nardi,
\newblock ``Robust spoken language understanding for house service robots,''
\newblock in {\em Proceedings of the 17th International Conference on
  Intelligent Text Processing and Computational Linguistics}, 2016, pp. 3--9.

\bibitem{L13}
P.~Lison and G.J.M. Kruijff,
\newblock ``{Salience-driven contextual priming of speech recognition for
  human-robot interaction},''
\newblock in {\em Proceedings of the 18th European Conference on Artificial
  Intelligence}, 2008.

\bibitem{sugiura09interspeech}
Komei Sugiura, Naoto Iwahashi, Hideki Kashioka, and Satoshi Nakamura,
\newblock ``Bayesian learning of confidence measure function for generation of
  utterances and motions in object manipulation dialogue task,''
\newblock in {\em Proceedings of Interspeech}, 2009, pp. 2483--2486.

\bibitem{sugiura15iros}
Komei Sugiura and Koji Zettsu,
\newblock ``Rospeex: A cloud robotics platform for human-robot spoken
  dialogues,''
\newblock in {\em Proc. {IEEE/RSJ IROS}}, 2015, pp. 6155--6160.

\bibitem{K25}
Ranjay Krishna, Yuke Zhu, Oliver Groth, Justin Johnson, Kenji Hata, Joshua
  Kravitz, Stephanie Chen, Yannis Kalantidis, Li-Jia Li, David~A Shamma,
  et~al.,
\newblock ``{Visual Genome: Connecting Language and Vision Using Crowdsourced
  Dense Image Annotations},'' arXiv:1602.07332, 2016.

\bibitem{odena2016conditional}
Augustus Odena, Christopher Olah, and Jonathon Shlens,
\newblock ``{Conditional Image Synthesis with Auxiliary Classifier GANs},''
  arXiv preprint arXiv:1610.09585, 2016.

\bibitem{L8}
Ronald~W. Langacker,
\newblock {\em Foundations of Cognitive Grammar: Theoretical Prerequisites},
\newblock Stanford Univ Pr, 1987.

\bibitem{I16}
Sergey Ioffe and Christian Szegedy,
\newblock ``{Batch Normalization: Accelerating Deep Network Training by
  Reducing Internal Covariate Shift},''
\newblock in {\em Prof. of ICML}, 2015, pp. 448--456.

\bibitem{H15}
Kaiming He, Xiangyu Zhang, Shaoqing Ren, and Jian Sun,
\newblock ``{Identity Mappings in Deep Residual Networks},''
\newblock in {\em Proc. of European Conference on Computer Vision}, 2016, pp.
  630--645.

\bibitem{L21}
Quoc Le and Tomas Mikolov,
\newblock ``{Distributed Representations of Sentences and Documents},''
\newblock in {\em Proc. of ICML}, 2014, pp. 1188--1196.

\bibitem{M17}
George~A. Miller et~al.,
\newblock ``{WordNet: a Lexical Database for English},''
\newblock {\em Communications of the ACM}, vol. 38, no. 11, pp. 39--41, 1995.

\bibitem{lin2014microsoft}
Tsung-Yi Lin, Michael Maire, Serge Belongie, James Hays, Pietro Perona, Deva
  Ramanan, Piotr Doll{\'a}r, and C~Lawrence Zitnick,
\newblock ``{Microsoft COCO: Common Objects in Context},''
\newblock in {\em European Conference on Computer Vision}, 2014, pp. 740--755.

\end{thebibliography}
